%% file: neurips_2026.tex
\documentclass{article}
\PassOptionsToPackage{numbers, compress}{natbib}
\usepackage[preprint]{neurips_2026}

\usepackage[utf8]{inputenc} 
\usepackage[T1]{fontenc}    
\usepackage{hyperref}       
\usepackage{url}            
\usepackage{booktabs}       
\usepackage{amsfonts}       
\usepackage{nicefrac}       
\usepackage{microtype}      
\usepackage{xcolor}         

\title{Implicit Compression Regularization: Concise Reasoning via Internal Shorter Distributions in RL Post-Training}

\usepackage{amsmath}
\usepackage{amssymb}
\usepackage{mathtools}
\usepackage{amsthm}
\renewcommand{\cite}{\citep}
\usepackage[T1]{fontenc}    
\usepackage{hyperref}       
\usepackage{url}            
\usepackage{booktabs}       
\usepackage{amsfonts}       
\usepackage{nicefrac}       
\usepackage{microtype}      
\usepackage{xcolor}         
\definecolor{citecolor}{HTML}{0071BC}
\definecolor{linkcolor}{HTML}{D32F2F}
\definecolor{cellcolor}{HTML}{E3F2FD}
\definecolor{red}{HTML}{D32F2F}
\definecolor{magenta}{HTML}{D81B60}

\usepackage{svg}
\svgsetup{inkscapearea=page}
\usepackage{amsmath}
\usepackage{amssymb}
\usepackage{mathtools}
\usepackage{amsthm}
\usepackage{amsmath, amssymb, amsthm}
\usepackage{amsfonts}
\usepackage[capitalize,noabbrev]{cleveref}

\usepackage{pgfplots}
\pgfplotsset{compat = newest}
\usepackage{multirow}
\usepackage{enumitem}
\usepackage{caption}
\usepackage{fancybox}
\usepackage{framed}
\usepackage{tcolorbox}
\usepackage{subcaption}
\usepackage{algorithm}
\usepackage{mathrsfs}
\usepackage{mathtools}
\usepackage{algorithmic}
\usepackage{tcolorbox}
\usepackage{listings}
\usepackage{makecell}
\usepackage{colortbl}
\usepackage{color}
\usepackage{wrapfig}
\usepackage{cancel}
\usepackage{soul,xcolor}
\usepackage{pifont}
\usepackage{tikz}
\usetikzlibrary{shapes, positioning, arrows.meta, calc, decorations.pathmorphing, quotes}
\tcbuselibrary{breakable}
\usepackage{hyperref}
\usepackage{makecell}
\usepackage{url}
\hypersetup{colorlinks=true, linkcolor=linkcolor, citecolor=citecolor,urlcolor=magenta}
\usepackage{booktabs}
\usepackage{hyperref}
\usepackage{url}
\usepackage{graphicx}
\usepackage{subcaption}
\usepackage{amsthm}
\newtheorem{claim}{Claim}
\usepackage{enumitem}
\usepackage{arydshln}
\setlist[itemize]{leftmargin=2em}
\setlist[enumerate]{leftmargin=2em}

\author{%
\begin{tabular}{c}
Chen Wang$^{1,2,*}$ \quad
Hexuan Deng$^{2,3}$ \quad
Yining Zhang$^{2,4}$ \quad
Yuchen Zhang$^{2,5}$ \\[0.25em]
Jionghao Bai$^{2,6}$ \quad
Zhaochun Li$^{2,7}$ \quad
Ge Lan$^{1,\dagger}$ \quad
Yue Wang$^{2,\dagger}$ \\[0.8em]
\normalfont $^{1}$College of Software, Nankai University \quad
\normalfont $^{2}$Zhongguancun Academy \\[0.25em]
\normalfont $^{3}$Harbin Institute of Technology \quad
\normalfont $^{4}$Institute of Automation, Chinese Academy of Sciences \\[0.25em]
\normalfont $^{5}$East China Normal University \quad
\normalfont $^{6}$Zhejiang University \quad
\normalfont $^{7}$Beijing Institute of Technology \\
\end{tabular}
}

\begin{document}

\maketitle

 \begingroup
 \renewcommand{\thefootnote}{}
 \footnotetext{$^*$Email: \texttt{s-wc25@bjzgca.edu.cn}.}
 \footnotetext{$^\dagger$ Correspondence to Ge Lan, email: \texttt{lange@nankai.edu.cn}.}
 \footnotetext{$^\dagger$ Correspondence to Yue Wang, email: \texttt{yuewang@bza.edu.cn}.}
 \endgroup
\begin{abstract}
Reinforcement learning with verifiable rewards improves LLM reasoning but often induces overthinking, where models generate unnecessarily long reasoning traces. Existing methods mainly rely on length penalties or early-exit strategies; however, the former may degrade accuracy and induce underthinking, whereas the latter assumes that substantial portions of reasoning traces can be safely truncated. To obtain a compression signal without these limitations, we revisit the training dynamics of existing compression methods. We observe that the length--accuracy correlation is initially negative but continually increases during compression, indicating that shorter responses are initially more likely to be correct but gradually lose this property as the policy moves toward underthinking. Based on this observation, we formalize overthinking: a negative correlation indicates an overthinking regime, while a positive one indicates underthinking. When overthinking, the shortest correct responses are shorter than the group-average response length in expectation, making them natural compression targets already present in on-policy rollouts. We therefore propose \emph{Implicit Compression Regularization} (ICR), an on-policy regularization method whose compression signal comes from a virtual shorter distribution induced by the shortest correct responses in rollout groups, guiding the policy toward concise yet correct trajectories. Training dynamics show that ICR maintains a better length--accuracy correlation during compression, indicating that short responses remain better aligned with correctness instead of drifting toward underthinking. Experiments on three reasoning backbones and multiple mathematical and knowledge-intensive benchmarks show that ICR consistently shortens responses while preserving or improving accuracy, achieving a stronger accuracy--length Pareto frontier.

\end{abstract}

\section{Introduction}

Large language models (LLMs) have achieved strong reasoning performance by scaling test-time computation through long chain-of-thought reasoning \citep{wei2022chain,guo2025deepseek,team2025kimi1_5}. Reinforcement learning with verifiable rewards (RLVR) further strengthens this capability by optimizing models with outcome-level correctness signals, enabling them to explore, reflect, and revise reasoning trajectories \citep{shao2024deepseekmath,guo2025deepseek,team2025kimi1_5}. However, longer reasoning is not always beneficial. During RL post-training, models may generate redundant intermediate steps, repeat self-reflections, or allocate excessive computation to questions that have already been solved \citep{su2025between,alomrani2025reasoning}. This phenomenon, commonly known as \emph{overthinking}, increases inference cost and can even hurt correctness by introducing spurious alternatives or unnecessary self-correction \citep{chen2024not,sui2025stop}. Therefore, an important problem is how to reduce redundant reasoning while preserving the reasoning capability acquired through RL.

Existing methods mainly address overthinking in two ways. The first category adds length penalties to the RL reward \citep{team2025kimi1_5,yu2025dapo, yi2025shorterbetter,liu2026length}. Although effective at reducing token usage, these methods make response length an explicit optimization target, which can degrade accuracy and make the policy prone to underthinking \citep{nohara2026optimal,yue2025does}. The second category uses early-exit or truncation-style strategies to stop reasoning once sufficient evidence is estimated to be available \citep{dai2025s,bin2025explore, cheng2025optimizing}. However, these methods rely on the assumption that large portions of reasoning traces are redundant and can be safely discarded, which may fail on harder, information-dense problems where later steps remain tightly coupled with correctness. These limitations motivate a different question: can we obtain a compression signal during on-policy training without length penalties or reasoning truncation?

Motivated by this question, we revisit the training dynamics of existing compression methods. By tuning the length coefficient, we observe that stronger length penalties shorten responses faster, but also accelerate accuracy degradation and worsen the accuracy--length Pareto frontier. More importantly, we find that the group-wise length--accuracy correlation starts negative but continually increases during compression. This indicates that shorter responses are initially more likely to be correct, suggesting the existence of safe compression opportunities within the current rollout distribution. However, as training proceeds, this property gradually vanishes, implying that the policy is pushed from removing redundancy toward underthinking. This suggests that compression itself is not inherently harmful, but directly optimizing shortness makes the policy exploit an easier optimization direction than improving correctness.
According to the observation, we formalize overthinking by the expected group-wise correlation between correctness and response length: a negative value indicates an overthinking regime, while a positive value indicates an underthinking regime. In the overthinking regime, correct responses are shorter than the group average in expectation, so the shortest correct samples naturally provide safe compression targets already present in on-policy rollouts. Based on this insight, we propose \emph{Implicit Compression Regularization} (ICR), an on-policy regularization method that extracts compression signals from these shortest correct samples. Instead of adding a handcrafted length-dependent reward or truncating reasoning traces, ICR uses the shortest correct responses within rollout groups to induce a virtual shorter distribution. This distribution guides the policy toward concise yet correct trajectories already discovered by its own rollouts. Training dynamics show that ICR maintains a better length--accuracy correlation during compression, indicating that short responses remain better aligned with correctness instead of drifting toward underthinking. Experiments across three reasoning backbones and multiple mathematical and knowledge-intensive benchmarks show that ICR consistently reduces response length while preserving or improving accuracy, achieves a stronger accuracy--length Pareto frontier, and remains compatible with mild length penalties when stronger compression is required.

Our contributions are summarized as follows:
\begin{itemize}
\item We reveal a key training dynamic behind compression: the group-wise length--accuracy correlation starts negative but continually increases, showing that short responses are initially more likely to be correct but gradually lose this advantage as compression moves toward underthinking.
\item We formalize overthinking by the expected group-wise correlation between correctness and response length: a negative value indicates an overthinking regime, while a positive value indicates an underthinking regime. 
\item We propose \emph{Implicit Compression Regularization} (ICR), an on-policy regularization method that extracts compression signals from the shortest correct samples within rollout groups, without introducing explicit length penalties or truncating reasoning traces.
\item We demonstrate across multiple backbones and benchmarks that ICR achieves accuracy-preserving compression, maintains a better length--accuracy correlation, and yields a stronger accuracy--length Pareto frontier. We further show that ICR is compatible with length penalties when stronger compression is required.\end{itemize}

\section{Related Work}
\label{sec:related_work}

\paragraph{RL post-training and GRPO.}
Reinforcement learning with verifiable rewards (RLVR) has become a central paradigm for improving the reasoning ability of LLMs. Given a query $q$ and a sampled response $o$, RLVR evaluates the response with a verifiable reward function $R(q,o)$ and maximizes
\begin{equation}
\mathcal{J}_{\rm RL}(\theta)=
\mathbb{E}_{q \sim P(Q),\, o \sim \pi_{\theta}}[R(q,o)].
\end{equation}
Among existing RLVR methods, Group Relative Policy Optimization (GRPO) is widely used for reasoning LLMs. For each query $q$, GRPO samples a group of $G$ responses $\{o_i\}_{i=1}^{G}$ from $\pi_{\theta_{\rm old}}$ and optimizes
\begin{equation}
\small
\mathcal{J}_{\rm GRPO}(\theta)=
\mathbb{E}_{q\sim P(Q),\,\{o_i\}_{i=1}^{G}\sim \pi_{\theta_{\rm old}}}
\left[
\frac{1}{G}\sum_{i=1}^{G}\frac{1}{|o_i|}\sum_{t=1}^{|o_i|}
\min\!\Big(
r_{i,t}(\theta)A_i,\,
\mathrm{clip}\!\big(r_{i,t}(\theta),1-\epsilon_{\text{low}},1+\epsilon_{\text{high}}\big)A_i
\Big)
\right],
\end{equation}
where $\epsilon_{\text{low}} = \epsilon_{\text{high}}=0.2$ and
\begin{equation}
r_{i,t}(\theta)=
\frac{\pi_{\theta}(o_{i,t}\mid q,o_{i,<t})}
{\pi_{\theta_{\rm old}}(o_{i,t}\mid q,o_{i,<t})}, \qquad
A_i=
\frac{
R_i-\mathrm{mean}(\{R_j\}_{j=1}^{G})
}{
\mathrm{std}(\{R_j\}_{j=1}^{G})
}.
\end{equation}
The group-normalized advantage reduces variance and stabilizes policy optimization. However, optimizing only final correctness often encourages long reasoning traces, leading to \emph{overthinking}, where models generate redundant reflections, repeated verification steps, or unnecessarily detailed derivations \citep{su2025between,alomrani2025reasoning,chen2024not,sui2025stop}. Such behavior increases inference cost and may even hurt correctness on hard or noisy problems \citep{fan2025missing,dang2025first,cuadron2025danger}.

\paragraph{Length penalties.}
A major line of work mitigates overthinking by adding a length-dependent term to the RL reward:
\begin{equation}
R_i = R_i^{\mathrm{corr}} + \lambda\, R_i^{\mathrm{len}},
\label{eq:total_reward_len}
\end{equation}
where $R_i^{\mathrm{corr}}$ is the correctness reward, $R_i^{\mathrm{len}}$ is the length reward, and $\lambda$ controls the compression strength. Since GRPO normalizes scalar rewards within each rollout group, the length term directly changes the relative advantage and participates in policy optimization. Existing length penalties can be summarized as follows:
\begin{itemize}
\item \textbf{LP-F.} Fixed-reference penalties use predefined length bounds or budgets to reward shorter responses, such as DAPO with a soft length shaping \citep{yu2025dapo}.
\item \textbf{LP-G.} Group-wise penalties normalize response lengths within the current rollout group and favor shorter samples relative to other responses, as in Kimi-k1.5 and so on \citep{team2025kimi1_5,arora2025training,liu2025learn,luo2026compress,liang2025deepcompress,hu2026smartthinker,su2025thinking}.
\end{itemize}
Although these methods can effectively reduce token usage, they explicitly couple response length with reward optimization. This makes training sensitive to coefficient tuning and may shift the policy toward superficial shortening rather than genuine reasoning improvement, causing underthinking or accuracy degradation \citep{liu2026length,nohara2026optimal,yue2025does}.
\begin{figure*}[t]
    \centering
    \includegraphics[width=0.9\linewidth]{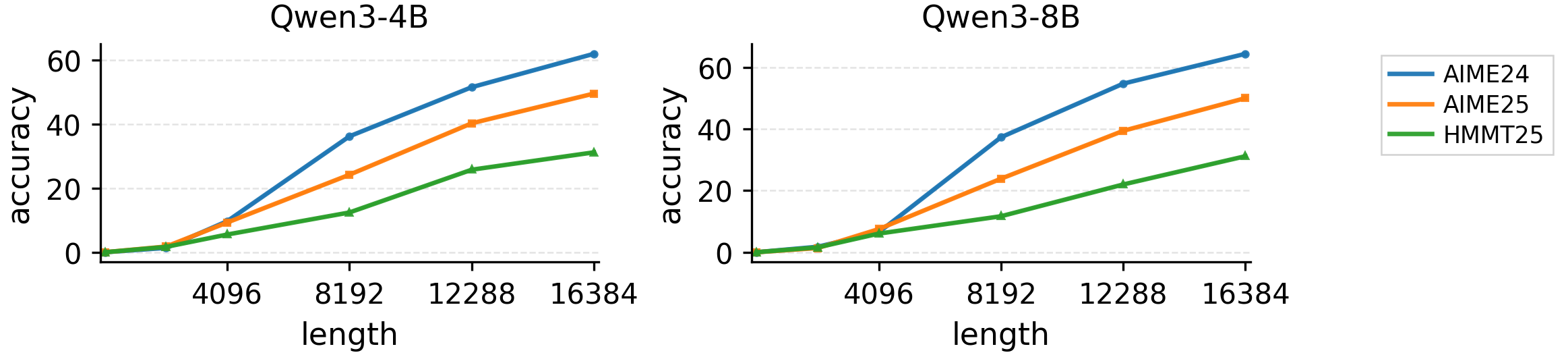}
    \caption{Accuracy under different maximum response lengths on mathematical reasoning benchmarks. Accuracy increases monotonically with the allowed reasoning length across both Qwen3-4B and Qwen3-8B, indicating that truncation can damage the quality of inference.}
    \label{fig:truncate}
\end{figure*}
\paragraph{Early exit and truncation.}
Another line of work reduces overthinking by constructing, selecting, or truncating reasoning trajectories. Some route queries between thinking and no-thinking modes \citep{zhang2025adaptthink,xu2025chain}. Early-exit and truncation-style methods instead shorten reasoning by stopping generation once a fixed budget is reached or once sufficient evidence is estimated to be available, using signals such as confidence, verification, or entropy \citep{dai2025s,bin2025explore,cheng2025optimizing}. Although these methods can reduce inference cost, they rely on the assumption that later reasoning steps are mostly redundant. As shown in Fig.~\ref{fig:truncate}, reasoning accuracy increases monotonically with the allowed response length on hard mathematical benchmarks, indicating that truncated steps often still contain useful reasoning computation. Therefore, early stopping may hurt performance when later steps are tightly coupled with correctness. In contrast, ICR does not discard reasoning trajectories by hard truncation or early exit, but extracts compression signals from concise correct samples already present in on-policy rollout groups.

\section{Method}
We begin by presenting an empirical observation that motivates our method. Although the length penalty is widely adopted to mitigate overthinking in RL post-training, we find that it inevitably incurs a performance loss. This observation reveals the limitation of existing reward-shaping strategies and motivates the method proposed in this section.

\begin{figure*}[t]
    \centering

    \begin{subfigure}[t]{\textwidth}
        \centering
        \includegraphics[width=0.32\linewidth]{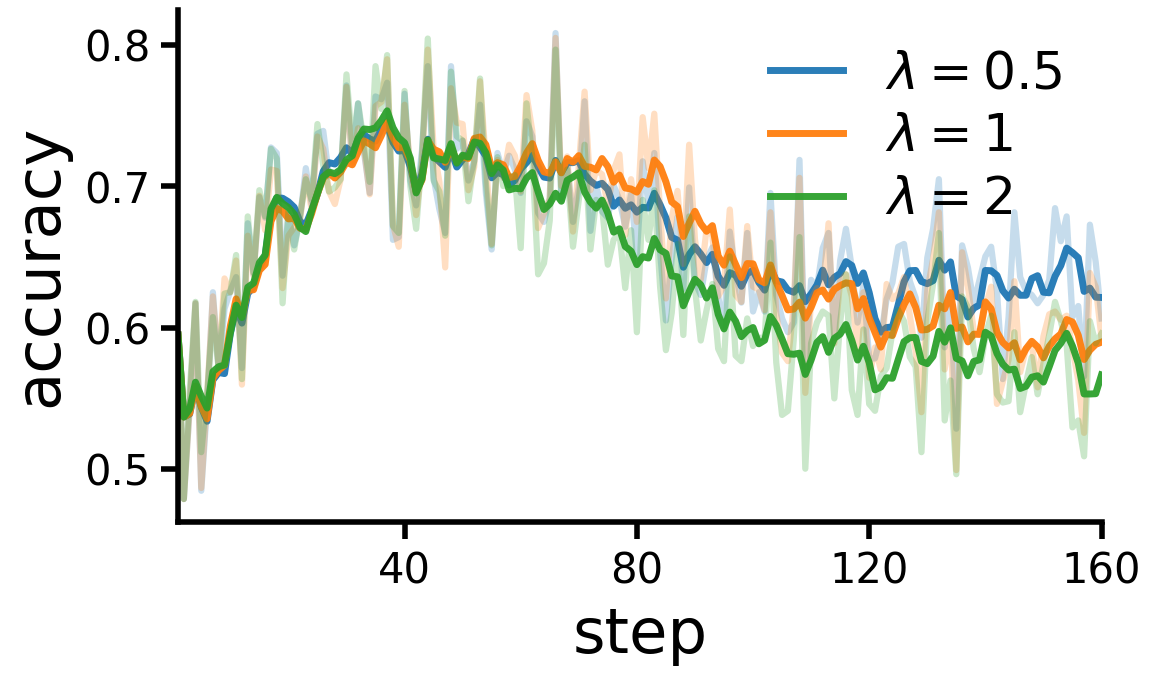}
        \includegraphics[width=0.32\linewidth]{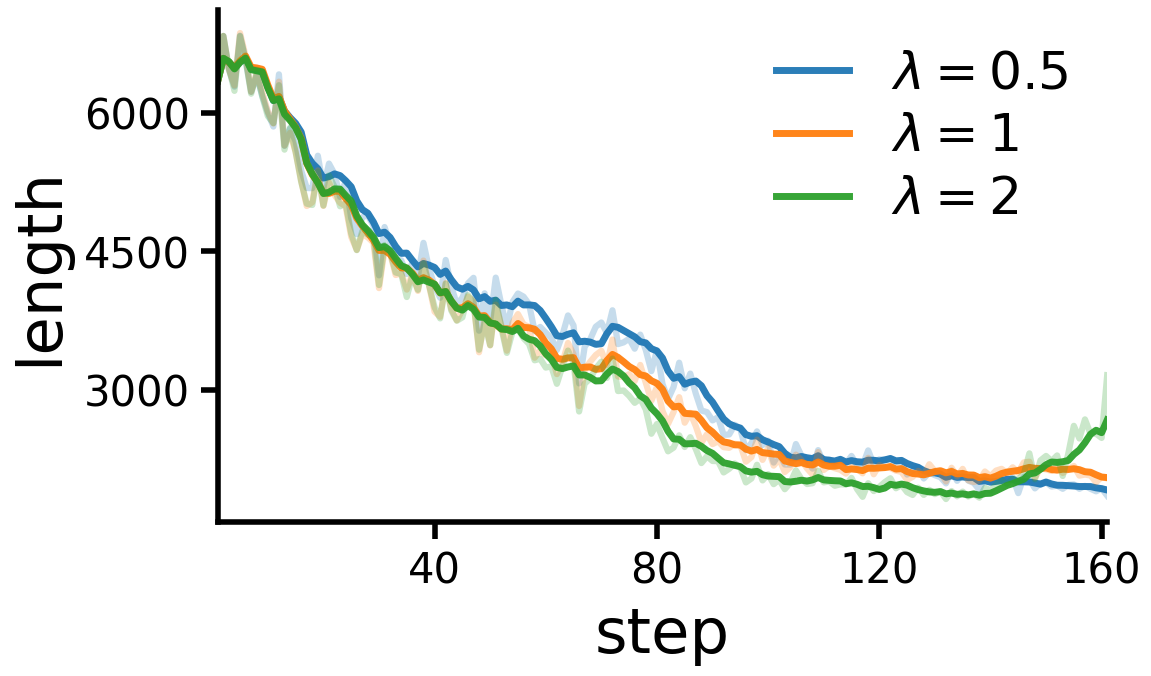}
        \includegraphics[width=0.32\linewidth]{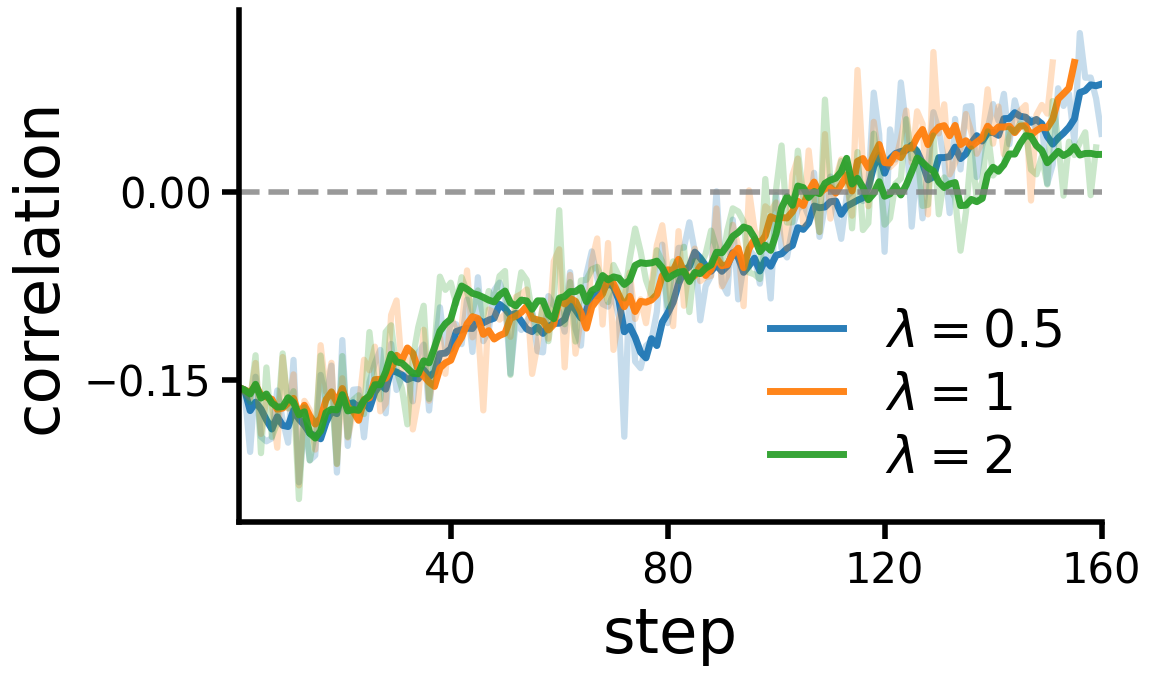}
        \caption{LP-F ($\ell_{\min}= 4096$, $\ell_{\max}= 8192$).}
    \end{subfigure}

    \vspace{0.5em}
    \begin{subfigure}[t]{\textwidth}
        \centering
        \includegraphics[width=0.32\linewidth]{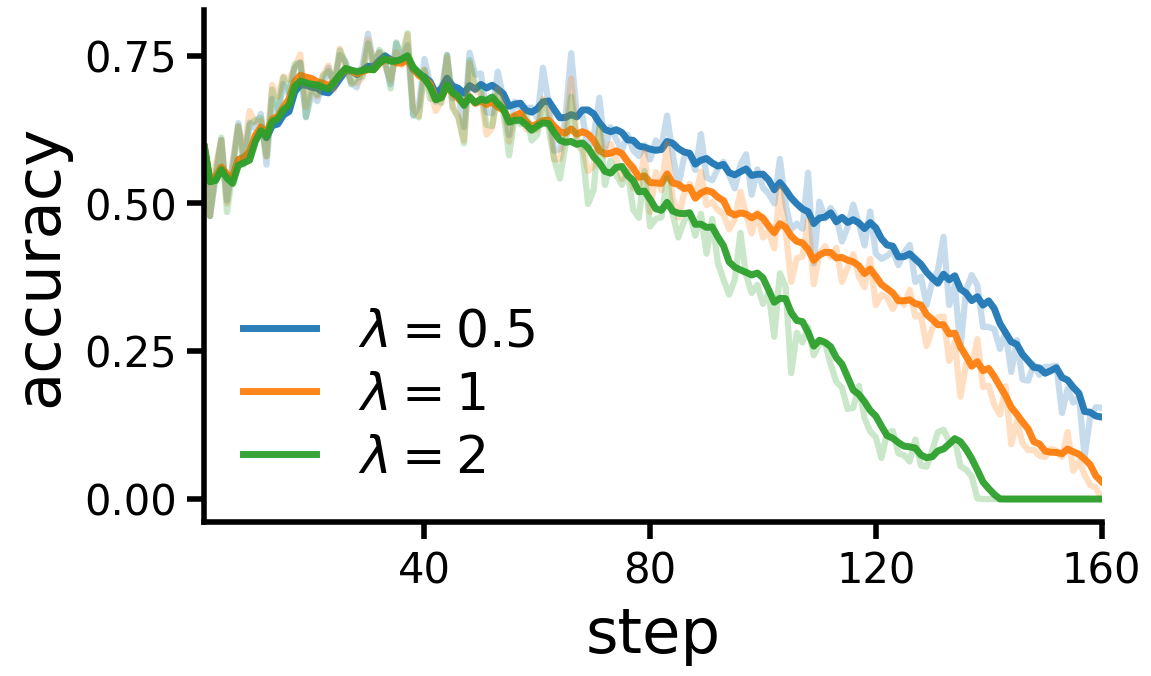}
        \includegraphics[width=0.32\linewidth]{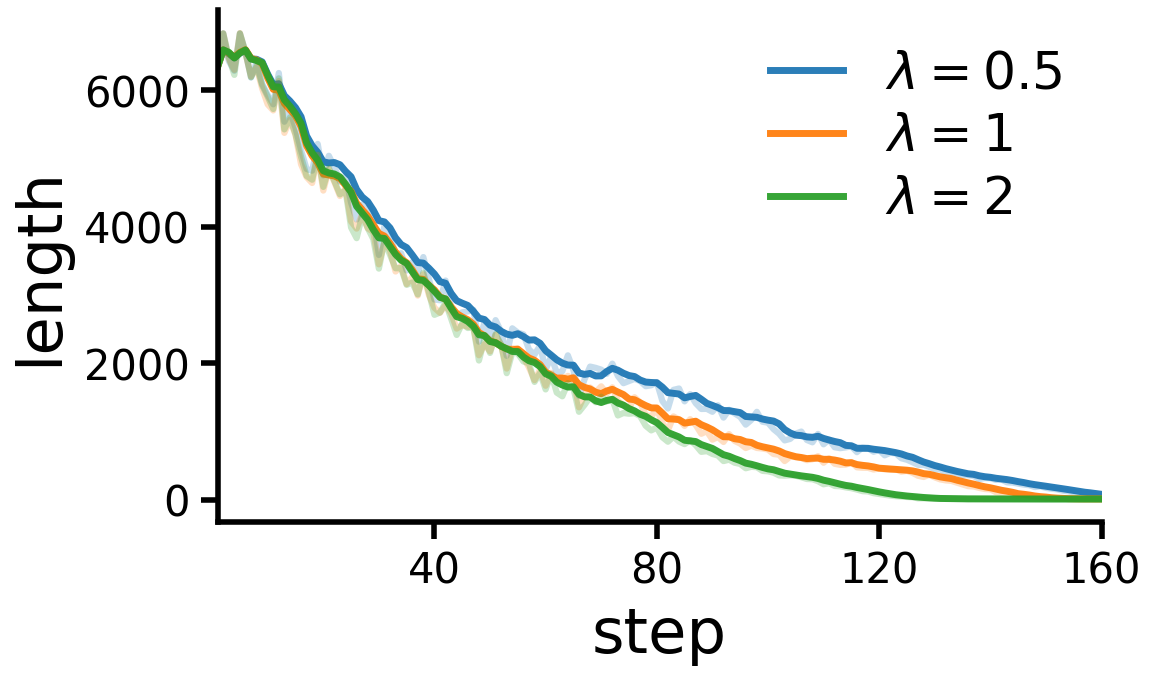}
        \includegraphics[width=0.32\linewidth]{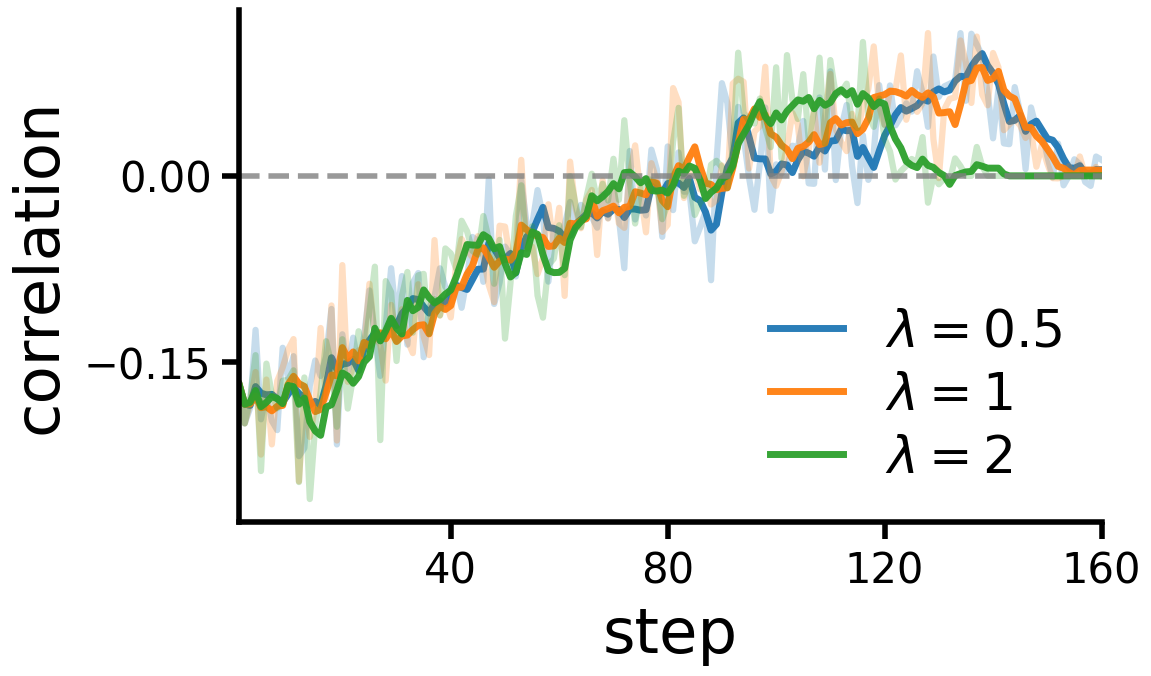}
        \caption{LP-G (Kimi-k1.5).}
    \end{subfigure}

    \caption{Coefficient-tuning results for the two length reward designs with $\lambda \in \{0.5,1,2\}$. In both cases, increasing the length coefficient shortens responses faster, but also accelerates accuracy degradation. The accuracy--length correlation starts negative and gradually moves toward zero, indicating that the initial compatibility between shorter responses and correctness weakens during training.}
    \label{fig:length_tuning_all}
\end{figure*}

\subsection{Observations}
\label{subsec:obs}

We tune the length coefficient $\lambda \in \{0.5,1,2\}$ in Eq.~\eqref{eq:total_reward_len} for both reward designs and track three quantities during training: training accuracy, average response length, and the correlation coefficient between response length and correctness reward. The correlation is computed within each rollout group and then averaged over all groups in the batch.

\textbf{Negative but weakening accuracy--length correlation.}
As shown in Fig.~\ref{fig:length_tuning_all}, for both penalty designs, the accuracy--length correlation is negative at the early stage of training. This means that, within the current rollout groups, shorter responses are often more likely to be correct, indicating that there is still room for safe compression. However, this negative correlation gradually moves toward zero as training proceeds. In other words, the advantage of shorter responses over longer ones becomes weaker during optimization. This trend suggests that length penalties exploit an initially available compression opportunity, but continuously applying the same pressure makes length reduction progressively less aligned with correctness.

\begin{figure*}[t]
    \centering
    \begin{subfigure}[t]{0.45\textwidth}
        \centering
        \includegraphics[width=\linewidth]{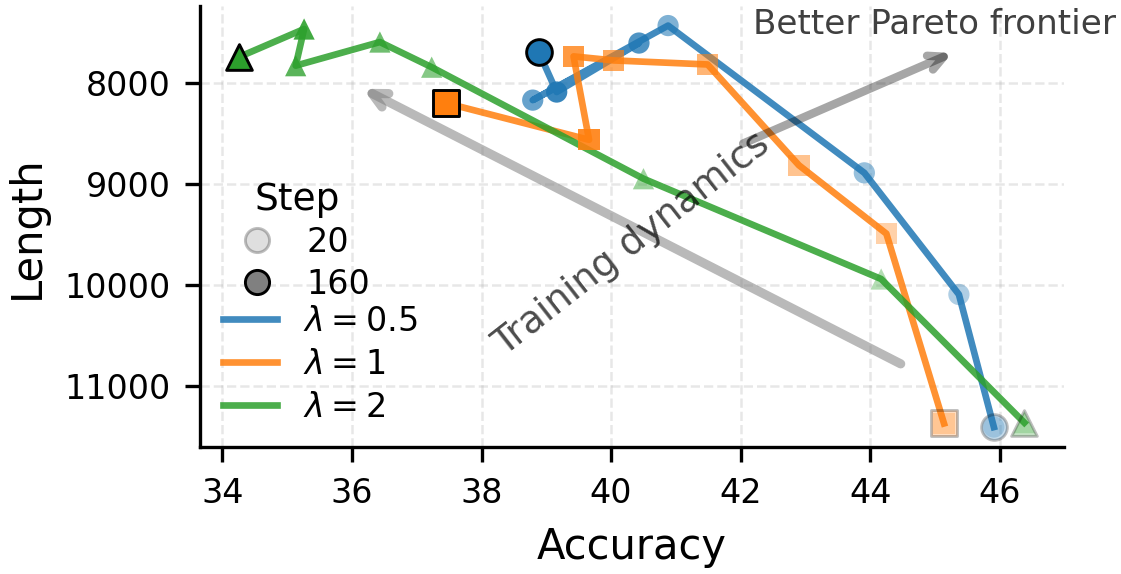}
        \caption{Accuracy--length trajectory of LP-F.}
    \end{subfigure}
    \hfill
    \begin{subfigure}[t]{0.45\textwidth}
        \centering
        \includegraphics[width=\linewidth]{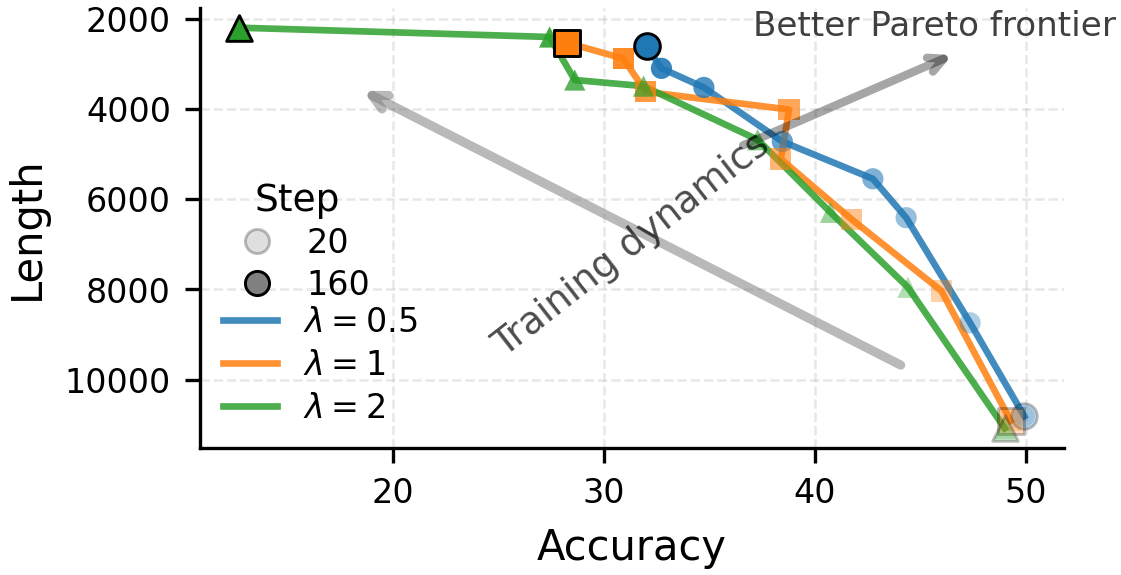}
        \caption{Accuracy--length trajectory of LP-G.}
    \end{subfigure}
    \caption{Accuracy--length trajectories on held-out mathematical competition benchmarks (AIME 24 \& AIME25 \& HMMT25). A larger length coefficient reaches shorter responses with fewer steps, but incurs a larger drop in reasoning accuracy.}
    \label{fig:length_acc_math}
\end{figure*}

\textbf{Accuracy--length Pareto frontier.}
As shown in Fig.~\ref{fig:length_tuning_all}, a larger length coefficient $\lambda$ makes the model shorten its responses faster on the training set, but the training accuracy also drops more rapidly. This indicates that stronger length pressure does not merely remove redundant tokens; it already begins to interfere with correctness optimization during training. Consistently, Fig.~\ref{fig:length_acc_math} shows that this degradation transfers to held-out mathematical competition benchmarks. Increasing $\lambda$ not only reduces performance, but also produces a worse accuracy--length Pareto frontier: for the same amount of length reduction, the model suffers a larger accuracy drop. Therefore, although a stronger length reward accelerates compression, it degrades the effective trade-off between reasoning budget and reasoning performance during both training and evaluation.

These observations suggest that explicit length penalties exploit the easier optimization direction of
shortening responses, which can gradually shift the policy from removing redundant reasoning toward
underthinking. This motivates a compression mechanism that uses the structure of on-policy rollouts
without directly rewarding shortness.

\subsection{Implicit Compression Regularization}
\label{subsec:method}

Based on this observation, we formalize the connection between overthinking and the group-wise correlation between accuracy and response length.

\begin{claim}[Overthinking / underthinking]
For a policy, a negative expected group-wise correlation between correctness and response length indicates an \emph{overthinking regime}, while a positive one indicates an \emph{underthinking regime}.
\end{claim}

A negative value means that correct responses are, on average, shorter within rollout groups, suggesting removable redundancy; a positive value suggests that correct responses are typically longer, so further compression may remove useful reasoning.

\begin{claim}[ICR]
For a policy, if the expected group-wise correlation between correctness and response length is negative, then the shortest correct responses are shorter than the group-average responses in expectation.
\end{claim}

This follows because negative expected correlation means that correct responses are shorter than the group average on average, and the shortest correct response is no longer than the average correct response. Therefore, shortest correct samples provide natural compression targets. Based on this, we propose \emph{Implicit Compression Regularization} (ICR), which extracts compression signals from shortest correct samples in on-policy rollout groups.

For a query $q$, let $\{o_i\}_{i=1}^{G}$ denote the rollout group sampled from $\pi_{\theta_{\rm old}}$. Each group already contains a natural compression target: the shortest correct sample in that group. We define the set of such samples as
\begin{equation}
\mathcal{S}(q)=
\arg\min_{i\in\{1,\dots,G\},\,R_i^{\rm corr}=1}|o_i|.
\label{eq:shortest_set}
\end{equation}

ICR augments RL with an additional preference toward these shortest correct responses:
\begin{equation}
\small
\begin{split}
\mathcal{J}_{\rm ICR}(\theta)
=&\,
\mathcal{J}_{\rm GRPO}(\theta)
+\alpha\,
\mathbb{E}_{q\sim P(Q),\,\{o_i\}_{i=1}^{G}\sim \pi_{\theta_{\rm old}}} \\ & \qquad\qquad\qquad
\Bigg[
\frac{1}{G}\sum_{i=1}^{G}
\frac{1}{|o_i|}\sum_{t=1}^{|o_i|}
\min\!\Big(
r_{i,t}(\theta),\,
\mathrm{clip}\!\big(
r_{i,t}(\theta),
1-\epsilon_{\rm low},
1+\epsilon_{\rm high}
\big)
\Big)
\mathbf{1}[o_i\in \mathcal{S}(q)]
\Bigg].
\\
\end{split}
\label{eq:j_icr}
\end{equation}

The regularizer term is activated only by the shortest correct samples in the current rollout group. Hence, ICR does not reward shortness itself; instead, it selectively amplifies trajectories that are both correct and already concise. This makes compression a consequence of reinforcing successful short reasoning paths, rather than an independent optimization target. This design gives ICR three key properties:
\begin{itemize}
\item \textbf{On-policy optimization.} ICR is fully on-policy. The selected samples come from the original rollout groups, and the compression signal in Eq.~\eqref{eq:j_icr} is obtained by a deterministic selection rule applied within each group. Therefore, ICR introduces neither an external policy nor an additional distributional mismatch, and avoids the risk of discarding useful later reasoning steps as in truncation-based methods.
\item \textbf{No explicit length penalty.} ICR does not add length penalties to the scalar reward. The primary optimization target remains reasoning correctness, while the compression signal is extracted from the group-wise structure of on-policy samples. In this way, ICR avoids directly reshaping the reward landscape toward the easier objective of shortening responses, which is the main source of degradation in length-penalty methods.
\item \textbf{No extra training cost.} ICR reuses the original rollout groups and only adds a lightweight group-wise selection step together with a simple advantage modification. Thus, ICR provides an effective compression signal without extra training cost.
\end{itemize}

\begin{figure*}[t]
    \centering
    \begin{subfigure}[t]{0.32\textwidth}
        \centering
        \includegraphics[width=\linewidth]{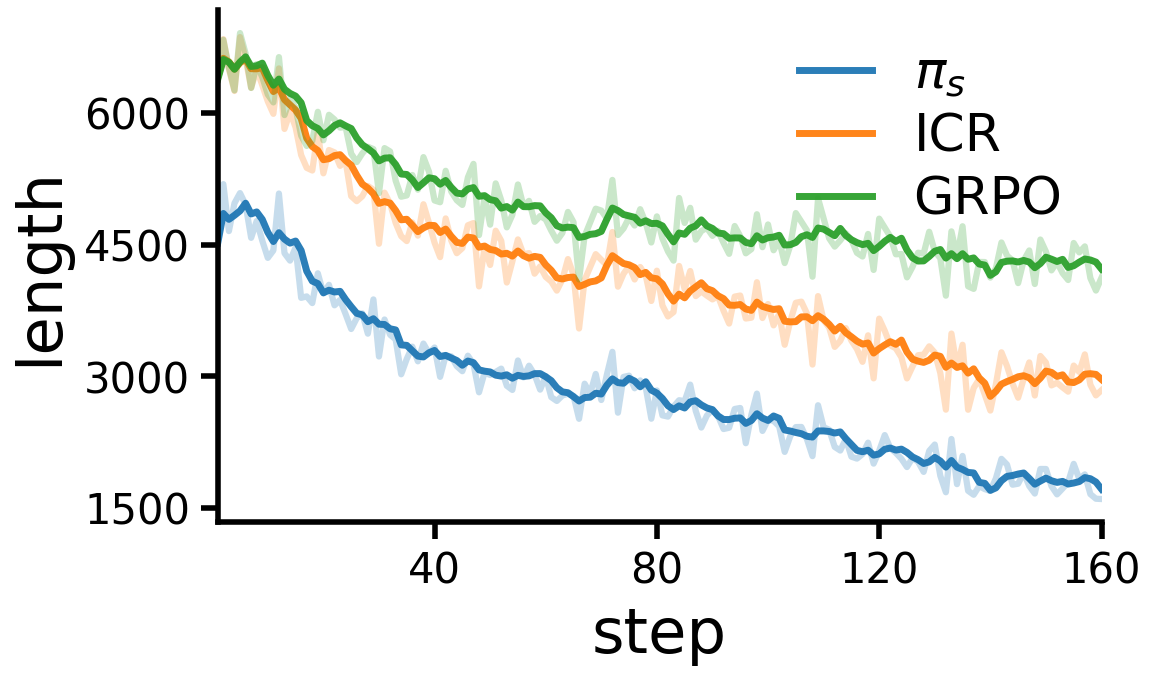}
        \caption{Qwen3-4B.}
    \end{subfigure}
    \hfill
    \begin{subfigure}[t]{0.32\textwidth}
        \centering
        \includegraphics[width=\linewidth]{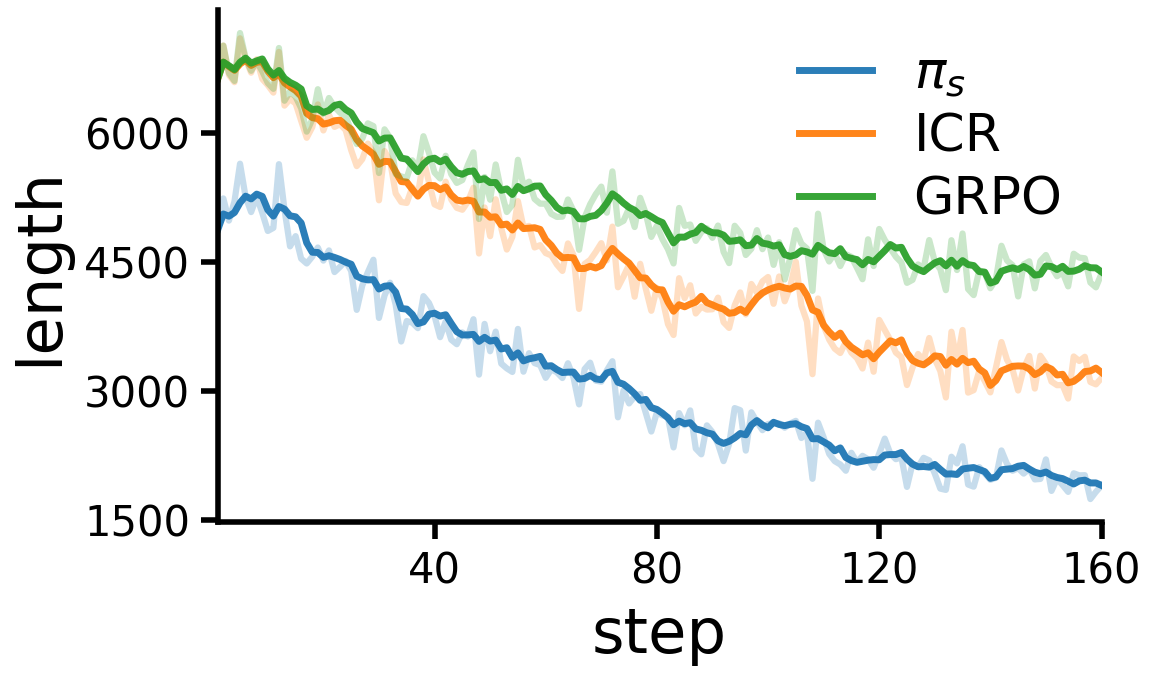}
        \caption{Qwen3-8B.}
    \end{subfigure}
    \hfill
    \begin{subfigure}[t]{0.32\textwidth}
        \centering
        \includegraphics[width=\linewidth]{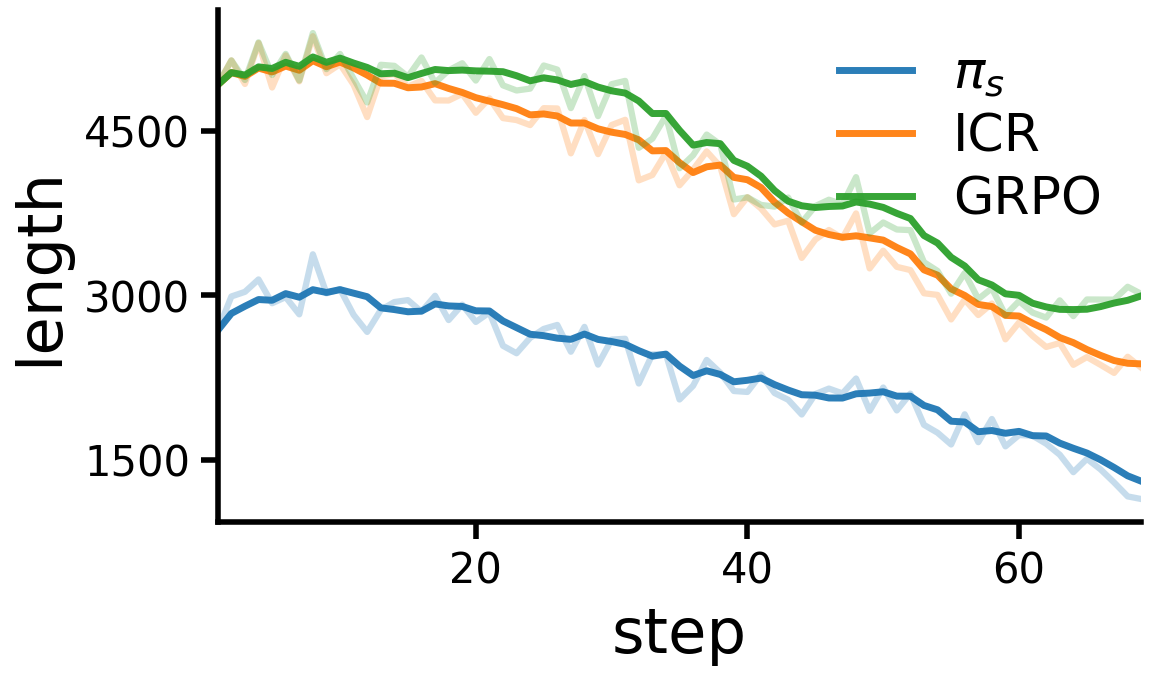}
        \caption{DSQW-7B.}
    \end{subfigure}
    \caption{Response length trajectories of $\pi_s$, ICR, and GRPO during training, showing that the shorter distribution $\pi_s$ guides the compression behavior of ICR.}
    \label{fig:pi_s_length}
\end{figure*}

The core mechanism of ICR is to turn the shortest correct samples in each rollout group into an implicit regularization signal. Since these samples are generated by the current policy and already achieve correct answers, they define a naturally ``shorter'' and successful region inside the on-policy distribution. From this perspective, the selected samples induce a virtual shorter distribution $\pi_s$, which retains the shortest successful trajectories from the original rollout distribution $\pi_\theta$. This induced distribution serves as the regularizer that guides compression: instead of optimizing a handcrafted length reward, ICR pulls the policy toward the concise successful trajectories that the policy itself has already discovered. As shown in Fig.~\ref{fig:pi_s_length}, $\pi_s$ consistently produces shorter responses than GRPO across different models, while ICR stays between them. This suggests that the shorter distribution $\pi_s$ effectively guides the policy toward concise reasoning, but does so softly enough to preserve the correctness-oriented optimization of RL.

\section{Experiments}
\begin{wrapfigure}{r}{0.45\linewidth}
	\centering
	\vspace{-0.8\baselineskip}
	\includegraphics[width=\linewidth]{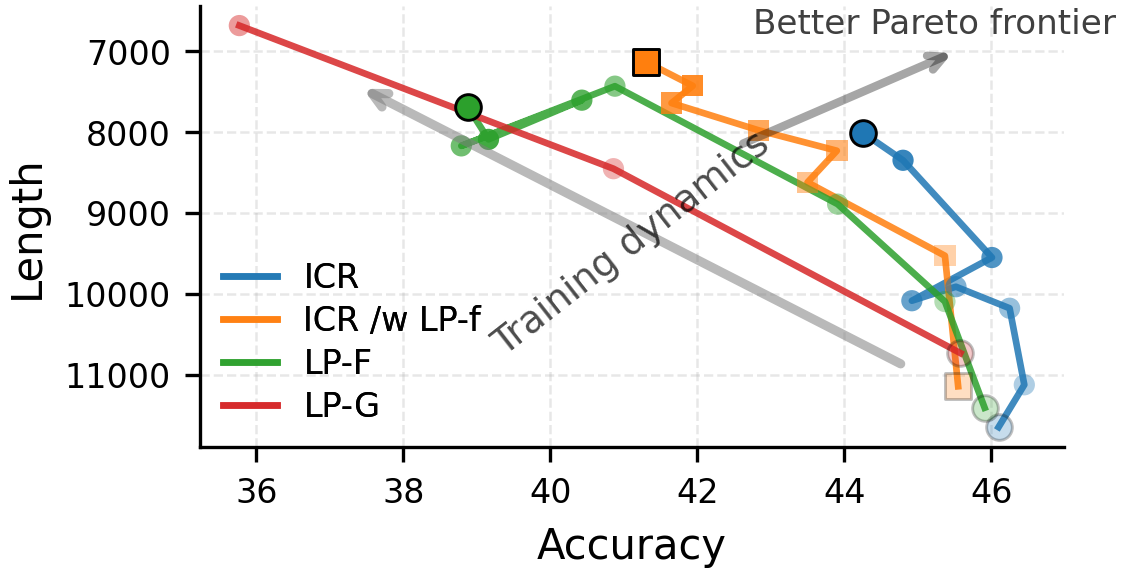}
	\vspace{-1.5\baselineskip}
	\caption{ICR and ICR w/ LP-F achieve a stronger accuracy–length Pareto frontier.}
	\label{fig:ICR-math}
\end{wrapfigure}
We evaluate ICR on three reasoning backbones, Qwen3-4B \citep{yang2025qwen3}, Qwen3-8B \citep{yang2025qwen3}, and DeepSeek-R1-Distill-Qwen-7B (DSQW-7B) \citep{guo2025deepseek}, using DAPO-17K \citep{yu2025dapo} for RL training. We compare ICR with representative methods from vanilla RL optimization, length-penalty methods, and inference-time compression, including GRPO \citep{shao2024deepseekmath}, LP-F \citep{yu2025dapo}, LP-G \citep{team2025kimi1_5}, ShorterBetter \citep{yi2025shorterbetter}, and LC-R1 \citep{cheng2025optimizing}. Evaluation covers both mathematical reasoning benchmarks and knowledge-intensive generalization benchmarks. Detailed baselines, datasets, and implementation settings are provided in Appendix~\ref{app:detail}.

\begin{table*}[t]
\caption{Main results on ICR and other length-aware reasoning compression methods.}
\label{tab:length_compression_results}
\centering
\resizebox{\textwidth}{!}{%
\renewcommand{\arraystretch}{1.35}
\begin{tabular}{
>{\raggedright\arraybackslash}m{3.3cm}
*{8}{>{\centering\arraybackslash}m{1.75cm}}
}
\toprule
\shortstack[l]{\textbf{Length}\\\textbf{Accuracy}} & AIME24 & AIME25 & HMMT25 & GSM8K & Math500 & AMC23 & Olympiad & \textbf{Average} \\
\midrule
\rowcolor{gray!10}
Qwen3-8B
& \makecell{12611 \\ 64.38}
& \makecell{13499 \\ 50.00}
& \makecell{14606 \\ 29.48}
& \makecell{1799 \\ 95.68}
& \makecell{4934 \\ 93.00}
& \makecell{7961 \\ 87.97}
& \makecell{9340 \\ 61.78}
& \makecell{9250 \\ 68.90} \\
+GRPO
& \makecell{\underline{8825} \\ \underline{66.35}}
& \makecell{\underline{8969} \\ \underline{52.70}}
& \makecell{\underline{10688} \\ \underline{31.67}}
& \makecell{\underline{1018} \\ \underline{96.05}}
& \makecell{\underline{2714} \\ \underline{93.60}}
& \makecell{\underline{4134} \\ \underline{89.06}}
& \makecell{\underline{5636} \\ \underline{65.03}}
& \makecell{\underline{5998} \\ \underline{70.64}} \\
\rowcolor{green!10}
+ICR
& \makecell{\textbf{7109} \\ \textbf{67.50}}
& \makecell{\textbf{8145} \\ \textbf{53.75}}
& \makecell{\textbf{9239} \\ \textbf{34.06}}
& \makecell{\textbf{713} \\ \textbf{96.13}}
& \makecell{\textbf{2208} \\ \textbf{95.00}}
& \makecell{\textbf{3645} \\ \textbf{89.53}}
& \makecell{\textbf{4869} \\ \textbf{66.22}}
& \makecell{\textbf{5133} \\ \textbf{71.74}} \\
\hdashline
\rowcolor{gray!10}
+LP-G (Kimi)
& \makecell{6282 \\ 54.58}
& \makecell{\underline{6424} \\ 40.41}
& \makecell{\underline{7609} \\ 24.37}
& \makecell{513 \\ 94.76}
& \makecell{\underline{1576} \\ 91.80}
& \makecell{2850 \\ 85.54}
& \makecell{\textbf{2517} \\ 59.55}
& \makecell{\underline{3967} \\ 64.43} \\
+ShorterBetter
& \makecell{\underline{6224} \\ 53.64}
& \makecell{6693 \\ 43.95}
& \makecell{7801 \\ 24.79}
& \makecell{\textbf{458} \\ 95.22}
& \makecell{\textbf{1504} \\ 91.80}
& \makecell{\underline{2577} \\ 86.64}
& \makecell{\underline{3267} \\ 61.03}
& \makecell{4075 \\ 65.30} \\
\rowcolor{gray!10}
+LC-R1
& \makecell{7035 \\ 56.87}
& \makecell{7873 \\ 42.39}
& \makecell{9289 \\ 23.95}
& \makecell{\underline{508} \\ 94.54}
& \makecell{1879 \\ 91.60}
& \makecell{2634 \\ 86.48}
& \makecell{4373 \\ 61.48}
& \makecell{4799 \\ 65.33} \\
+LP-F
& \makecell{6603 \\ \underline{60.72}}
& \makecell{6479 \\ \underline{44.68}}
& \makecell{7630 \\ \underline{28.33}}
& \makecell{618 \\ \underline{95.52}}
& \makecell{1757 \\ \underline{92.60}}
& \makecell{3271 \\ \underline{88.35}}
& \makecell{3817 \\ \textbf{63.11}}
& \makecell{4311 \\ \underline{67.62}} \\
\rowcolor{green!10}
+ICR /w LP-F
& \makecell{\textbf{6171} \\ \textbf{62.60}}
& \makecell{\textbf{5878} \\ \textbf{45.10}}
& \makecell{\textbf{6929} \\ \textbf{29.68}}
& \makecell{533 \\ \textbf{95.52}}
& \makecell{1684 \\ \textbf{94.40}}
& \makecell{\textbf{2292} \\ \textbf{88.75}}
& \makecell{3382 \\ \underline{62.96}}
& \makecell{\textbf{3838} \\ \textbf{68.43}} \\

\midrule
\rowcolor{gray!10}
Qwen3-4B
& \makecell{12354 \\ \underline{61.88}}
& \makecell{12759 \\ 49.48}
& \makecell{13920 \\ 31.15}
& \makecell{1602 \\ 94.84}
& \makecell{4681 \\ 91.60}
& \makecell{7360 \\ 87.50}
& \makecell{8990 \\ 62.37}
& \makecell{8809 \\ \underline{68.40}} \\
+GRPO
& \makecell{\underline{8636} \\ 61.87}
& \makecell{\underline{9087} \\ \underline{50.52}}
& \makecell{\underline{9427} \\ \underline{31.67}}
& \makecell{\underline{915} \\ \textbf{96.05}}
& \makecell{\underline{2533} \\ \underline{93.60}}
& \makecell{\underline{4199} \\ \underline{89.06}}
& \makecell{\underline{5027} \\ \underline{63.40}}
& \makecell{\underline{5689} \\ 69.45} \\
\rowcolor{green!10}
+ICR
& \makecell{\textbf{6588} \\ \textbf{66.97}}
& \makecell{\textbf{7274} \\ \textbf{52.60}}
& \makecell{\textbf{8442} \\ \textbf{32.60}}
& \makecell{\textbf{612} \\ \underline{95.14}}
& \makecell{\textbf{1884} \\ \textbf{93.60}}
& \makecell{\textbf{3878} \\ \textbf{93.04}}
& \makecell{\textbf{4333} \\ \textbf{65.33}}
& \makecell{\textbf{4716} \\ \textbf{71.33}} \\
\hdashline
\rowcolor{gray!10}
+LP-G (Kimi)
& \makecell{\textbf{4444} \\ 43.75}
& \makecell{\underline{5867} \\ 37.29}
& \makecell{\textbf{5394} \\ 20.52}
& \makecell{437 \\ 93.33}
& \makecell{\textbf{1008} \\ 88.60}
& \makecell{\textbf{1677} \\ 82.26}
& \makecell{\textbf{2122} \\ 56.59}
& \makecell{\textbf{2993} \\ 60.33} \\
+ShorterBetter
& \makecell{\underline{4891} \\ 46.25}
& \makecell{\textbf{5825} \\ 34.68}
& \makecell{\underline{6481} \\ 20.72}
& \makecell{\textbf{336} \\ 93.47}
& \makecell{\underline{1189} \\ 90.40}
& \makecell{\underline{1799} \\ 82.89}
& \makecell{\underline{2590} \\ 57.77}
& \makecell{\underline{3302} \\ 60.88} \\
\rowcolor{gray!10}
+LC-R1
& \makecell{6882 \\ 41.35}
& \makecell{7134 \\ 36.52}
& \makecell{7611 \\ 23.43}
& \makecell{\underline{362} \\ 93.70}
& \makecell{1724 \\ 88.80}
& \makecell{2890 \\ 81.64}
& \makecell{4338 \\ 55.85}
& \makecell{4420 \\ 60.18} \\
+LP-F
& \makecell{6674 \\ \underline{54.79}}
& \makecell{7350 \\ \underline{44.47}}
& \makecell{8535 \\ \underline{26.14}}
& \makecell{777 \\ \underline{94.54}}
& \makecell{2127 \\ \underline{91.80}}
& \makecell{3250 \\ \textbf{89.06}}
& \makecell{4269 \\ \underline{63.11}}
& \makecell{4712 \\ \underline{66.27}} \\
\rowcolor{green!10}
+ICR /w LP-F
& \makecell{5783 \\ \textbf{55.83}}
& \makecell{6622 \\ \textbf{46.97}}
& \makecell{7644 \\ \textbf{30.31}}
& \makecell{549 \\ \textbf{94.76}}
& \makecell{1695 \\ \textbf{92.40}}
& \makecell{2699 \\ \underline{88.13}}
& \makecell{3716 \\ \textbf{64.14}}
& \makecell{4101 \\ \textbf{67.51}} \\

\midrule

\rowcolor{gray!10}
DSQW-7B
& \makecell{10836 \\ 41.25}
& \makecell{11639 \\ 28.54}
& \makecell{\underline{12831} \\ 14.69}
& \makecell{\underline{1091} \\ 90.60}
& \makecell{3302 \\ 87.20}
& \makecell{6222 \\ 77.50}
& \makecell{7448 \\ 49.19}
& \makecell{7624 \\ 55.57} \\
+GRPO
& \makecell{\underline{10119} \\ \underline{47.70}}
& \makecell{\underline{11352} \\ \underline{35.72}}
& \makecell{13443 \\ \underline{19.79}}
& \makecell{1185 \\ \underline{92.34}}
& \makecell{\underline{3132} \\ \underline{90.20}}
& \makecell{\underline{5226} \\ \underline{84.06}}
& \makecell{\underline{7229} \\ \underline{54.37}}
& \makecell{\underline{7384} \\ \underline{60.60}} \\
\rowcolor{green!10}
+ICR
& \makecell{\textbf{8399} \\ \textbf{50.52}}
& \makecell{\textbf{9980} \\ \textbf{36.56}}
& \makecell{\textbf{12123} \\ \textbf{19.79}}
& \makecell{\textbf{1006} \\ \textbf{92.49}}
& \makecell{\textbf{2791} \\ \textbf{91.00}}
& \makecell{\textbf{4159} \\ \textbf{85.31}}
& \makecell{\textbf{6677} \\ \textbf{55.85}}
& \makecell{\textbf{6448} \\ \textbf{61.65}} \\
\hdashline
\rowcolor{gray!10}
+LP-G (Kimi)
& \makecell{8559 \\ 46.04}
& \makecell{10618 \\ \underline{35.10}}
& \makecell{9644 \\ \underline{20.94}}
& \makecell{596 \\ 91.21}
& \makecell{2216 \\ 90.60}
& \makecell{3392 \\ 85.47}
& \makecell{5521 \\ 54.67}
& \makecell{5792 \\ 60.58} \\
+ShorterBetter
& \makecell{\textbf{7543} \\ \underline{47.91}}
& \makecell{\underline{8738} \\ 34.37}
& \makecell{\underline{9105} \\ 20.00}
& \makecell{\textbf{516} \\ 91.81}
& \makecell{\textbf{1910} \\ 91.40}
& \makecell{\underline{3172} \\ 86.64}
& \makecell{\textbf{5021} \\ 55.70}
& \makecell{\underline{5144} \\ 61.12} \\
\rowcolor{gray!10}
+LC-R1
& \makecell{8460 \\ 46.97}
& \makecell{9714 \\ 34.79}
& \makecell{11499 \\ 20.10}
& \makecell{\underline{554} \\ 92.34}
& \makecell{2165 \\ 91.40}
& \makecell{3618 \\ 87.18}
& \makecell{5572 \\ 56.14}
& \makecell{5940 \\ 61.27} \\
+LP-F
& \makecell{8203 \\ 47.70}
& \makecell{9121 \\ 34.27}
& \makecell{9571 \\ 20.52}
& \makecell{734 \\ \underline{92.41}}
& \makecell{2313 \\ \underline{91.80}}
& \makecell{3779 \\ \underline{87.50}}
& \makecell{5374 \\ \underline{57.03}}
& \makecell{5585 \\ \underline{61.60}} \\
\rowcolor{green!10}
+ICR /w LP-F
& \makecell{\underline{7714} \\ \textbf{49.89}}
& \makecell{\textbf{8537} \\ \textbf{35.20}}
& \makecell{\textbf{8657} \\ \textbf{21.45}}
& \makecell{616 \\ \textbf{93.32}}
& \makecell{\underline{2078} \\ \textbf{93.00}}
& \makecell{\textbf{2698} \\ \textbf{87.89}}
& \makecell{\underline{5106} \\ \textbf{57.62}}
& \makecell{\textbf{5058} \\ \textbf{62.62}} \\

\bottomrule
\end{tabular}%
}
\end{table*}

\begin{table*}[t]
\caption{Generalization results on knowledge-intensive benchmarks, including ARC-Challenge, MMLU-Pro, and SuperGPQA.}
\label{tab:general_bench_results}
\centering
\resizebox{\textwidth}{!}{%
\renewcommand{\arraystretch}{1.35}
\begin{tabular}{
>{\raggedright\arraybackslash}m{2.5cm}
*{3}{>{\centering\arraybackslash}m{1.75cm}}
*{3}{>{\centering\arraybackslash}m{1.75cm}}
*{3}{>{\centering\arraybackslash}m{1.75cm}}
}
\toprule
\multirow{2}{*}{\textbf{Methods}}
& \multicolumn{3}{c}{\textbf{Qwen3-4B}}
& \multicolumn{3}{c}{\textbf{Qwen3-8B}}
& \multicolumn{3}{c}{\textbf{DSQW-7B}} \\
\cline{2-10}
& ARC & MMLU & S-GPQA
& ARC & MMLU & S-GPQA
& ARC & MMLU & S-GPQA \\
\midrule

\rowcolor{gray!10}
Base
& \makecell{1231 \\ \underline{93.31}}
& \makecell{2312 \\ \underline{72.86}}
& \makecell{6858 \\ 37.20}
& \makecell{1398 \\ \underline{92.64}}
& \makecell{2631 \\ 76.07}
& \makecell{7262 \\ \textbf{44.20}}
& \makecell{\underline{1321} \\ 78.60}
& \makecell{2993 \\ 56.61}
& \makecell{6238 \\ 30.00} \\
+GRPO
& \makecell{\underline{886} \\ 92.57}
& \makecell{\underline{1268} \\ 72.32}
& \makecell{\underline{4143} \\ \underline{41.00}}
& \makecell{\underline{981} \\ 92.57}
& \makecell{\underline{1486} \\ \underline{79.10}}
& \makecell{\underline{4022} \\ 42.60}
& \makecell{1386 \\ \underline{81.60}}
& \makecell{\underline{2515} \\ \underline{57.32}}
& \makecell{\underline{5452} \\ \underline{31.80}} \\
\rowcolor{green!10}
+ICR
& \makecell{\textbf{622} \\ \textbf{93.64}}
& \makecell{\textbf{830} \\ \textbf{73.39}}
& \makecell{\textbf{3082} \\ \textbf{41.20}}
& \makecell{\textbf{804} \\ \textbf{93.31}}
& \makecell{\textbf{1195} \\ \textbf{79.46}}
& \makecell{\textbf{3498} \\ \underline{44.00}}
& \makecell{\textbf{1177} \\ \textbf{82.61}}
& \makecell{\textbf{2136} \\ \textbf{59.82}}
& \makecell{\textbf{4635} \\ \textbf{33.20}} \\
\hdashline
\rowcolor{gray!10}
+LP-G (Kimi)
& \makecell{\underline{464} \\ 92.30}
& \makecell{\underline{720} \\ 71.25}
& \makecell{\textbf{1300} \\ 37.20}
& \makecell{646 \\ 92.64}
& \makecell{\underline{751} \\ 75.17}
& \makecell{\underline{2280} \\ 39.40}
& \makecell{855 \\ 79.60}
& \makecell{1877 \\ \underline{58.39}}
& \makecell{3277 \\ 31.80} \\
+ShorterBetter
& \makecell{\textbf{418} \\ 92.97}
& \makecell{\textbf{566} \\ 71.25}
& \makecell{\underline{2009} \\ 34.80}
& \makecell{\underline{614} \\ 91.97}
& \makecell{\textbf{725} \\ 76.78}
& \makecell{\textbf{2269} \\ 40.80}
& \makecell{\underline{707} \\ 81.27}
& \makecell{\underline{1676} \\ 57.85}
& \makecell{\underline{2967} \\ 31.60} \\
\rowcolor{gray!10}
+LC-R1
& \makecell{505 \\ 91.63}
& \makecell{805 \\ \underline{72.14}}
& \makecell{2884 \\ 36.20}
& \makecell{701 \\ 92.64}
& \makecell{946 \\ 76.42}
& \makecell{3079 \\ 40.40}
& \makecell{878 \\ \underline{83.27}}
& \makecell{1693 \\ 57.67}
& \makecell{3832 \\ 32.60} \\
\rowcolor{gray!10}
+LP-F
& \makecell{732 \\ \underline{93.64}}
& \makecell{1000 \\ 71.60}
& \makecell{3189 \\ \underline{38.20}}
& \makecell{699 \\ \underline{93.64}}
& \makecell{811 \\ \textbf{79.28}}
& \makecell{3392 \\ \underline{41.60}}
& \makecell{792 \\ 82.60}
& \makecell{1688 \\ 57.85}
& \makecell{3390 \\ \underline{33.40}} \\
\rowcolor{green!10}
+ICR /w LP-F
& \makecell{608 \\ \textbf{93.97}}
& \makecell{752 \\ \textbf{73.39}}
& \makecell{2392 \\ \textbf{39.60}}
& \makecell{\textbf{602} \\ \textbf{93.97}}
& \makecell{754 \\ \underline{78.03}}
& \makecell{2312 \\ \textbf{41.60}}
& \makecell{\textbf{603} \\ \textbf{84.28}}
& \makecell{\textbf{1666} \\ \textbf{59.28}}
& \makecell{\textbf{2957} \\ \textbf{35.20}} \\

\bottomrule
\end{tabular}%
}
\end{table*}

\subsection{Main results}

\begin{figure*}[t]
    \centering
    \includegraphics[width=0.32\linewidth]{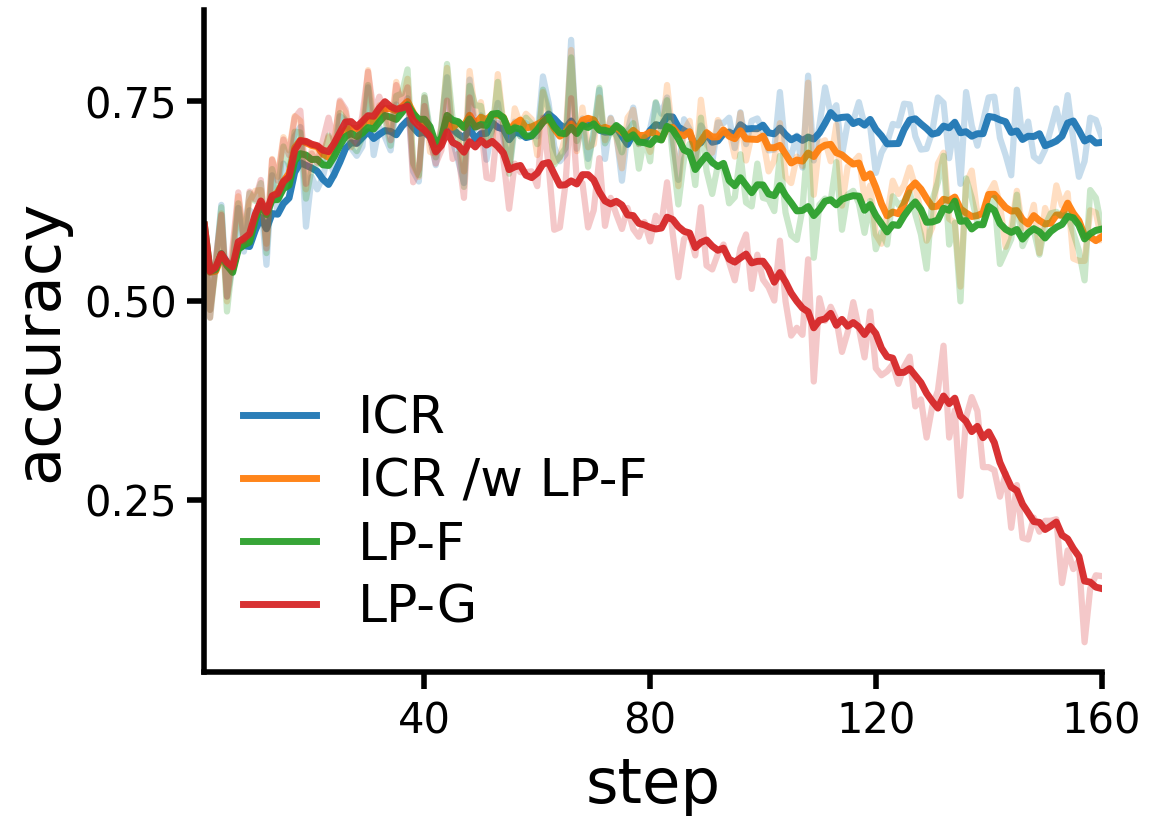}
    \includegraphics[width=0.32\linewidth]{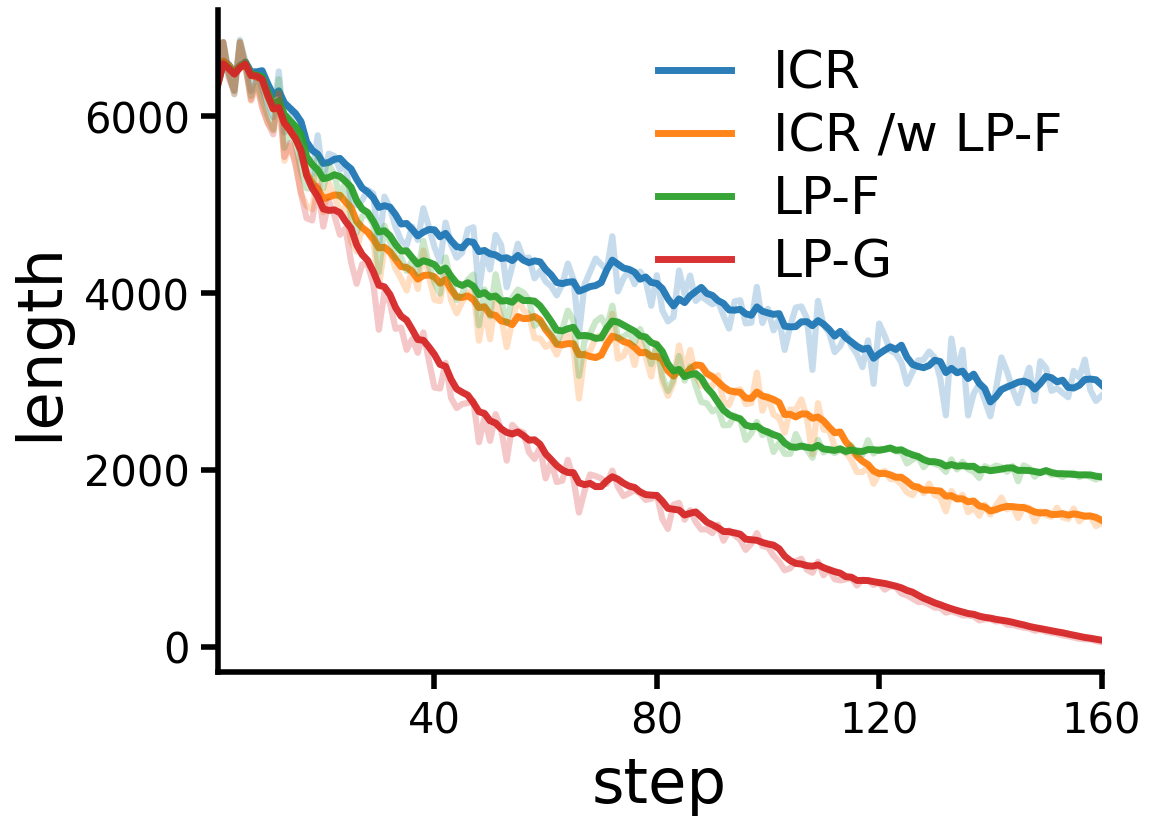}
    \includegraphics[width=0.32\linewidth]{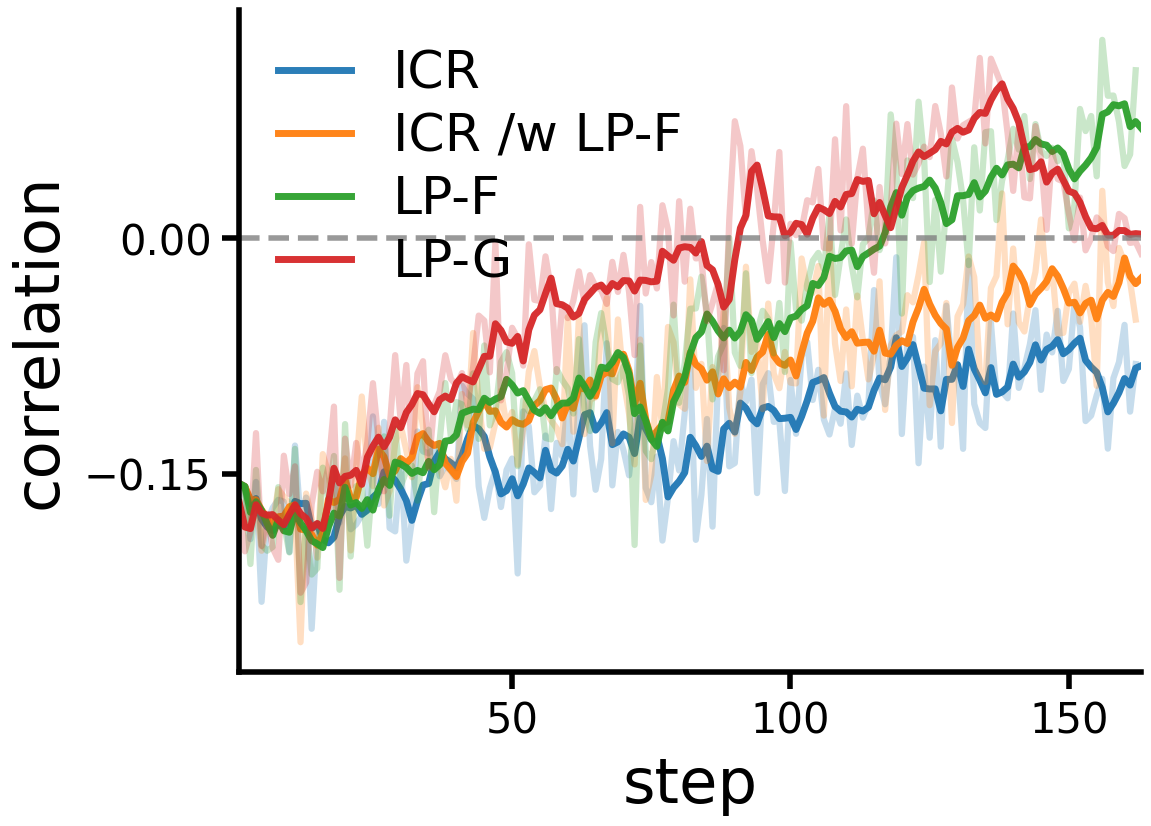}
    \caption{Training dynamics of ICR and ICR /w LP-F on mathematical reasoning benchmarks. From left to right: training accuracy, average response length, and the group-wise accuracy--length correlation averaged over the batch.}
    \label{fig:ICR-alc}
\end{figure*}

\textbf{Accuracy-preserving compression}
Tables~\ref{tab:length_compression_results} and \ref{tab:general_bench_results} report the main results on mathematical reasoning and knowledge-intensive generalization benchmarks, respectively. Overall, ICR consistently achieves accuracy-preserving compression. Compared with GRPO, ICR produces shorter responses while further improving accuracy, showing that overthinking can be reduced without sacrificing reasoning quality. The training dynamics in Fig.~\ref{fig:ICR-alc} provide further evidence. ICR maintains the most stable accuracy curve throughout training, whereas length-penalty baselines, especially LP-G, exhibit stronger late-stage degradation. More importantly, ICR keeps a more negative accuracy--length correlation during training, indicating that shorter samples remain more aligned with correctness. This suggests that ICR preserves the quality of concise responses during compression, rather than pushing the model toward underthinking. Moreover, Fig.~\ref{fig:ICR-math} shows that ICR achieves the most favorable accuracy--length Pareto frontier, indicating that its compression is not only effective but also better aligned with reasoning performance.

\textbf{Compatibility with length penalty.}
ICR provides a soft implicit compression signal that preserves accuracy but usually compresses more moderately than aggressive length-reward baselines. Therefore, an important question is whether ICR can complement length penalties when stronger compression is required. We find that ICR is highly compatible with length penalties. When combined with LP-F, ICR further strengthens the compression effect of the soft length reward while maintaining the correctness-oriented preference induced by the shortest correct samples. As shown in Table~\ref{tab:length_compression_results} and Fig.~\ref{fig:ICR-math}, ICR /w LP-F achieves the best overall performance and the most favorable accuracy--length Pareto frontier among all baselines. Moreover, compared with LP-F alone, ICR /w LP-F produces even shorter responses with higher accuracy. The training dynamics in Fig.~\ref{fig:ICR-alc} further show that ICR /w LP-F keeps a more negative accuracy--length correlation than LP-F, indicating that ICR strengthens the quality of short samples while enhancing compression. This suggests that ICR does not conflict with length penalties; instead, it acts as a complementary on-policy regularizer that guides the model toward concise yet correct reasoning trajectories.

\textbf{Generalization.}
Another important finding is the strong generalization of ICR. The gains of ICR are consistent across all three backbones, including Qwen3-4B, Qwen3-8B, and DSQW-7B, and transfer well from mathematical competition benchmarks to knowledge-intensive benchmarks such as ARC-Challenge, MMLU-Pro, and SuperGPQA. These results suggest that the benefit of ICR does not depend on a specific model family or benchmark type, but reflects a more general improvement in the reasoning compression trade-off.

\textbf{Baseline behaviors.}
Compared with the base models, GRPO already provides a favorable starting point by shortening responses while maintaining, and often improving, reasoning accuracy. In contrast, length-penalty methods achieve substantially stronger compression, but their gains in efficiency are often accompanied by noticeable performance degradation, especially on mathematical reasoning benchmarks. The early-exit baseline LC-R1 performs relatively well on DSQW-7B and on knowledge-intensive generalization benchmarks. However, its accuracy drop becomes much more pronounced on more reasoning-intensive base models, such as Qwen3, and on harder competition-level mathematical benchmarks. This suggests that early-exit methods rely on the existence of long but largely redundant reasoning prefixes that can be safely truncated. When reasoning becomes more information-dense, however, later steps are less redundant and more tightly coupled to correctness, making early exit much less effective as a compression strategy. In contrast, ICR does not depend on reward penalty or premature truncation, but instead extracts compression signals from successful short trajectories already contained in on-policy rollouts. This allows ICR to achieve a more favorable balance between compression and accuracy across both reasoning-intensive and general-domain settings.

\subsection{Ablations}

We ablate three key designs of ICR on Qwen3-4B: selecting all samples, selecting negative samples, and using only the regularizer without the GRPO objective. Table~\ref{tab:ablation_results} reports the final results, and the curves below show the corresponding training dynamics.

\textbf{All samples.}
When all samples enter $\mathcal{S}(q)$, the regularizer loses its preference for shorter samples. The length curve confirms that compression is largely weakened, and the average length increases from 4716 to 6541 tokens. Accuracy also drops slightly from 71.33 to 70.64, suggesting that applying the unnormalized regularization signal to all samples introduces higher-variance updates.

\textbf{Negative samples.}
Allowing negative samples in $\mathcal{S}(q)$ brings only limited compression, with the average length increasing to 5888 tokens. The training curve also stays consistently above ICR. This shows that under-reasoned short samples are not useful compression targets; their additional optimization signal offsets the benefit of reinforcing concise correct samples, leading to a slight accuracy drop.

\textbf{Only regularizer.}
Using only the regularizer achieves strong compression, with an average length of 4815 tokens and the fastest length reduction in training. However, accuracy drops clearly from 71.33 to 68.44. This indicates that the shortest-correct regularizer can induce concise responses, but cannot preserve reasoning accuracy alone. Therefore, the normalized GRPO objective is crucial to ICR, and the compression regularizer should remain an auxiliary term.

\begin{table*}[t]
\caption{Ablation results of different variants of ICR.}
\label{tab:ablation_results}
\centering
\resizebox{\textwidth}{!}{%
\renewcommand{\arraystretch}{1.35}
\begin{tabular}{
>{\raggedright\arraybackslash}m{4.4cm}
*{7}{>{\centering\arraybackslash}m{1.65cm}}
>{\raggedright\arraybackslash}m{2.5cm}
}
\toprule
\textbf{Qwen3-4B}
& AIME24 & AIME25 & HMMT25 & GSM8K & Math500 & AMC23 & Olympiad & \makecell{\textbf{Average}} \\
\midrule
\rowcolor{red!10}
All samples in $\mathcal{S}(q)$
& \makecell{8332 \\ 64.89}
& \makecell{10652 \\ 51.25}
& \makecell{11713 \\ 33.12}
& \makecell{1127 \\ 95.22}
& \makecell{2963 \\ 92.80}
& \makecell{4749 \\ 93.97}
& \makecell{6254 \\ 63.25}
& \makecell{\textbf{6541 \textcolor{red}{(+1825)}} \\ 70.64 \textbf{\textcolor{red}{(-0.69)}}} \\
\rowcolor{red!10}
Negative samples in $\mathcal{S}(q)$
& \makecell{8910 \\ 66.77}
& \makecell{9113 \\ 52.50}
& \makecell{10086 \\ 32.70}
& \makecell{974 \\ 94.61}
& \makecell{2625 \\ 92.80}
& \makecell{4006 \\ 92.64}
& \makecell{5499 \\ 63.85}
& \makecell{\textbf{5888 \textcolor{red}{(+1172)}}\\ 70.84 \textbf{\textcolor{red}{(-0.49)}}} \\

\rowcolor{red!10}
Only regularizer
& \makecell{7859 \\ 61.08}
& \makecell{7906 \\ 48.75}
& \makecell{8310 \\ 29.58}
& \makecell{614 \\ 94.16}
& \makecell{1725 \\ 91.60}
& \makecell{2860 \\ 91.56}
& \makecell{4433 \\ 62.37}
& \makecell{\textbf{4815 \textcolor{red}{(+99)}}\\ 68.44 \textbf{\textcolor{red}{(-2.89)}}} \\

\bottomrule
\end{tabular}%
}\\
\vspace{0.5em}
\includegraphics[width=1\linewidth]{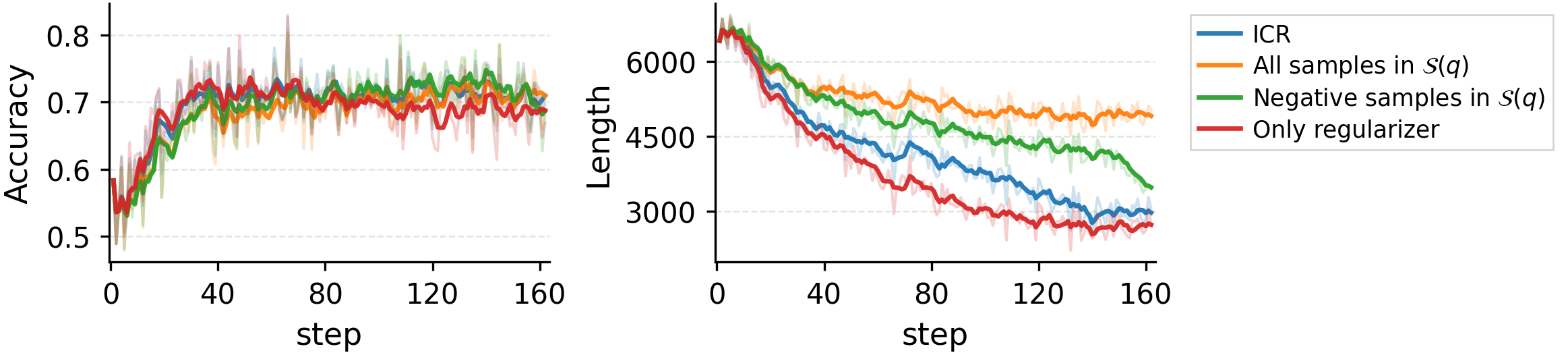}
\vspace{-2em}
\end{table*}

\section{Conclusion}
In this work, we studied overthinking in RL post-training from the perspective of the relationship between response length and correctness. We formalized overthinking using the expected group-wise correlation between correctness and response length, where negative correlation indicates that shorter responses are more likely to be correct, while positive correlation suggests a drift toward underthinking. Based on this view, we proposed \emph{Implicit Compression Regularization} (ICR), which extracts compression signals from the shortest correct samples already present in on-policy rollout groups. By inducing a virtual shorter distribution, ICR regularizes the policy toward concise successful trajectories without adding a handcrafted length reward. Experiments across multiple backbones and benchmarks demonstrate that ICR achieves accuracy-preserving compression, generalizes beyond mathematical reasoning, and remains compatible with length penalties for stronger compression.  

As discussed in Limitations (Sec.~\ref{sec:limitations} in appendix), future work should therefore move beyond designing stronger compression heuristics and further investigate the underlying relationship between response length and reasoning accuracy. A key direction is to understand what determines the accuracy--length Pareto frontier during RL post-training, including how model capacity, task difficulty, rollout diversity, reward structure, and optimization dynamics shape the attainable trade-off. Such analysis may help distinguish removable redundancy from reasoning steps that are genuinely necessary for correctness. Ultimately, this line of research points toward lossless reasoning compression, where models can reduce unnecessary computation while preserving the full reasoning accuracy of long responses.

\bibliographystyle{unsrt}
\bibliography{reference}


\appendix

\section{Limitations \label{sec:limitations}}

A limitation of this work lies in the granularity of our definition of overthinking and underthinking. To the best of our knowledge, we make the first attempt to distinguish these two regimes using the length--accuracy correlation during RL post-training: a negative value indicates an overthinking regime, while a positive value indicates an underthinking regime. This definition provides a simple and useful batch-level diagnostic for identifying whether shorter responses are still aligned with correctness. However, it is also coarse-grained. Since the correlation is computed over rollout groups and averaged at the batch level, it cannot determine whether a specific response, reasoning step, or query is truly redundant. A batch may also contain both overthinking and underthinking examples, which are only summarized by a single aggregate statistic. Therefore, this definition should be viewed as an initial operational boundary rather than a fine-grained characterization of reasoning redundancy. Developing instance-level or step-level criteria for distinguishing removable redundancy from necessary reasoning remains an important direction for future work.

Moreover, ICR provides a soft compression signal rather than an explicit length-control mechanism. This property is desirable for preserving accuracy, since ICR does not directly optimize a handcrafted notion of shortness. Our experiments show that ICR can achieve a more favorable accuracy--length Pareto frontier by guiding the policy toward concise yet correct trajectories. However, this also means that ICR does not directly enforce a target response length or a strict inference budget. When aggressive compression is required, ICR may need to be combined with length penalties or other budget-aware mechanisms. Nevertheless, this compatibility also reflects a limitation: achieving stronger or controllable compression may still require reintroducing explicit length-based constraints. Future work should explore how to achieve stronger compression without explicit length rewards while further preserving reasoning accuracy.

\section{Detailed implementation \label{app:detail}}

\begin{table}[t]
\centering
\caption{Detailed implementation for all experiments.}
\label{tab:implementation}
\renewcommand{\arraystretch}{1.15}
\begin{tabular}{l l}
\hline
\textbf{Hardware} & 8$\times$ A800 GPUs (40GB) \\
\hline
\textbf{Training Framework} & EasyR1 / VeRL \\
\textbf{Training Dataset} & DAPO-17K \\
\hline
\textbf{RL Settings:} & \\
\quad Maximum response length & 8192 \\
\quad Batch size $|B|$ & 128 \\
\quad Mini batch size & 64 \\
\quad Rollout group size $G$ & 8 \\
\quad Sampling temperature & 1.0 \\
\quad Learning rate & $1\times 10^{-6}$ \\
\quad Clip range $\epsilon=\epsilon_{\rm low}=\epsilon_{\rm high}$ & 0.2 \\
\quad Reward type & Binary correctness reward \\
\hline
\textbf{ICR Settings:} & \\
\quad Selection rule & Shortest correct sample(s) in each group \\
\quad Empty correct set & No ICR regularization for this group \\
\quad ICR coefficient & $\alpha=\alpha_0|B|/|\mathcal{S}(q)|$, $\alpha_0=0.5$ \\
\hline
\textbf{Length-Penalty Settings:} & \\
\quad LP-F coefficient $\lambda$ & 0.5 \\
\quad LP-F lower bound $\ell_{\min}$ & 4096 \\
\quad LP-F upper bound $\ell_{\max}$ & 8192 \\
\quad LP-G setting & Group-wise min-max normalization \\
\quad ICR /w LP-F & ICR combined with the same LP-F setting \\
\hline
\textbf{Evaluation Settings:} & \\
\quad Maximum response length & 16384 \\
\quad Top P & 0.95 \\
\quad Temperature & 0.1 \\
\hline
\end{tabular}
\end{table}

\paragraph{Experimental setup.}
We implement ICR in the EasyR1 and VeRL frameworks \citep{zheng2025easyr1,sheng2025hybridflow} and follow the default EasyR1 setup unless otherwise specified. Experiments are conducted on Qwen3-4B, Qwen3-8B, and DeepSeek-R1-Distill-Qwen-7B (DSQW-7B), using DAPO-17K \citep{yu2025dapo} for training. Baselines are chosen from three categories: vanilla RL optimization, length-penalty methods, and inference-time compression. Specifically, we compare with GRPO \citep{shao2024deepseekmath}, LP-F from DAPO with $\ell_{\min}=4096$ and $\lambda=0.5$ \citep{yu2025dapo}, LP-G from Kimi-K1.5 \citep{team2025kimi1_5}, ShorterBetter, and LC-R1.

\paragraph{Evaluation and implementation details.}
We evaluate in-domain mathematical reasoning on AIME24 \cite{balunovic2025matharena}, AIME25 \cite{balunovic2025matharena}, HMMT25 \cite{balunovic2025matharena}, AMC23 \citep{lightman2023lets}, GSM8K \citep{cobbe2021gsm8k}, MATH500 \citep{lightman2023lets} and Olympiad \citep{lightman2023lets}, and further evaluate out-of-domain generalization on ARC-Challenge \citep{clark2018think}, MMLU-Pro \citep{wang2024mmlu}, and SuperGPQA \citep{du2025supergpqa}. All experiments are conducted on 8 A800 GPUs, and the full training configuration is listed in Table~\ref{tab:implementation}. For ICR, the shortest correct samples are selected within each rollout group. If a group contains no correct response, the ICR regularizer is not applied to that group. If multiple correct responses share the minimum length, all of them are selected, and the regularization coefficient is normalized by the number of selected samples.


\newpage
\input{checklist.tex}

\end{document}

%% file: checklist.tex
\section*{NeurIPS Paper Checklist}

\begin{enumerate}

\item {\bf Claims}
    \item[] Question: Do the main claims made in the abstract and introduction accurately reflect the paper's contributions and scope?
    \item[] Answer: \answerYes{} 
    \item[] Justification: We have mentioned in the abstract.

\item {\bf Limitations}
    \item[] Question: Does the paper discuss the limitations of the work performed by the authors?
    \item[] Answer: \answerYes{} 
    \item[] Justification: We have discussed it in Section Limitation (in appendix).

\item {\bf Theory assumptions and proofs}
    \item[] Question: For each theoretical result, does the paper provide the full set of assumptions and a complete (and correct) proof?
    \item[] Answer: \answerNA{}.
    \item[] Justification: There are no theoretical results in this paper.

    \item {\bf Experimental result reproducibility}
    \item[] Question: Does the paper fully disclose all the information needed to reproduce the main experimental results of the paper to the extent that it affects the main claims and/or conclusions of the paper (regardless of whether the code and data are provided or not)?
    \item[] Answer: \answerYes{} 
    \item[] Justification: We use open-sourced data and models, which are easy to reproduce.

\item {\bf Open access to data and code}
    \item[] Question: Does the paper provide open access to the data and code, with sufficient instructions to faithfully reproduce the main experimental results, as described in supplemental material?
    \item[] Answer: \answerYes{} 
    \item[] Justification: We provide codes in our supplemental material. The data and models we used are all open-sourced.

\item {\bf Experimental setting/details}
    \item[] Question: Does the paper specify all the training and test details (e.g., data splits, hyperparameters, how they were chosen, type of optimizer) necessary to understand the results?
    \item[] Answer: \answerYes{} 
    \item[] Justification: We provide in Detailed implementation in appendix.

\item {\bf Experiment statistical significance}
    \item[] Question: Does the paper report error bars suitably and correctly defined or other appropriate information about the statistical significance of the experiments?
    \item[] Answer: \answerNo{} 
    \item[] Justification: Experiments on LLMs are too expensive to run for many times.

\item {\bf Experiments compute resources}
    \item[] Question: For each experiment, does the paper provide sufficient information on the computer resources (type of compute workers, memory, time of execution) needed to reproduce the experiments?
    \item[] Answer: \answerYes{} 
    \item[] Justification: We provide in our appendix.
    
\item {\bf Code of ethics}
    \item[] Question: Does the research conducted in the paper conform, in every respect, with the NeurIPS Code of Ethics \url{https://neurips.cc/public/EthicsGuidelines}?
    \item[] Answer: \answerYes{} 
    \item[] Justification: We follow NeurIPS Code of Ethics.

\item {\bf Broader impacts}
    \item[] Question: Does the paper discuss both potential positive societal impacts and negative societal impacts of the work performed?
    \item[] Answer: \answerNA{} 
    \item[] Justification:  There is no societal impact of the work performed.
    
\item {\bf Safeguards}
    \item[] Question: Does the paper describe safeguards that have been put in place for responsible release of data or models that have a high risk for misuse (e.g., pre-trained language models, image generators, or scraped datasets)?
    \item[] Answer: \answerNA{} 
    \item[] Justification: The paper poses no such risks

\item {\bf Licenses for existing assets}
    \item[] Question: Are the creators or original owners of assets (e.g., code, data, models), used in the paper, properly credited and are the license and terms of use explicitly mentioned and properly respected?
    \item[] Answer: \answerYes{} 
    \item[] Justification:  We follow the license and terms of use in our experiments.

\item {\bf New assets}
    \item[] Question: Are new assets introduced in the paper well documented and is the documentation provided alongside the assets?
    \item[] Answer: \answerNA{} 
    \item[] Justification: The paper does not release new assets

\item {\bf Crowdsourcing and research with human subjects}
    \item[] Question: For crowdsourcing experiments and research with human subjects, does the paper include the full text of instructions given to participants and screenshots, if applicable, as well as details about compensation (if any)? 
    \item[] Answer: \answerNA{} 
    \item[] Justification:  The paper does not involve crowdsourcing nor research with human subjects.

\item {\bf Institutional review board (IRB) approvals or equivalent for research with human subjects}
    \item[] Question: Does the paper describe potential risks incurred by study participants, whether such risks were disclosed to the subjects, and whether Institutional Review Board (IRB) approvals (or an equivalent approval/review based on the requirements of your country or institution) were obtained?
    \item[] Answer: \answerNA{} 
    \item[] Justification: The paper does not involve crowdsourcing nor research with human subjects.

\item {\bf Declaration of LLM usage}
    \item[] Question: Does the paper describe the usage of LLMs if it is an important, original, or non-standard component of the core methods in this research? Note that if the LLM is used only for writing, editing, or formatting purposes and does \emph{not} impact the core methodology, scientific rigor, or originality of the research, declaration is not required.
    \item[] Answer: \answerNA{} 
    \item[] Justification: The core method development in this research does not involve LLMs as any important, original, or non-standard components.

\end{enumerate}